\documentclass[letterpaper, 10 pt, journal, twoside]{IEEEtran}
\IEEEoverridecommandlockouts 

\usepackage{amsmath,amsfonts}
\usepackage{array}
\usepackage[caption=false,font=normalsize,labelfont=sf,textfont=sf]{subfig}
\usepackage{textcomp}
\usepackage{stfloats}
\usepackage{url}
\usepackage{verbatim}
\usepackage{graphicx}
\usepackage[table, dvipsnames]{xcolor} 
\usepackage{multirow}
\usepackage{mathtools}
\let\labelindent\relax
\usepackage{enumitem} 
\usepackage{booktabs}
\graphicspath{{figures/}}
\usepackage{caption}
\usepackage{tabularx}
\usepackage{threeparttable}
\usepackage{makecell}
\usepackage{transparent}

\def\BibTeX{{\rm B\kern-.05em{\sc i\kern-.025em b}\kern-.08em
    T\kern-.1667em\lower.7ex\hbox{E}\kern-.125emX}}
\usepackage{balance}

\usepackage[pagewise]{lineno}

\usepackage{booktabs}
\usepackage[noadjust]{cite}
\nocite{*}  

\makeatletter
\let\NAT@parse\undefined
\makeatother

\usepackage{hyperref}
\hypersetup{
    colorlinks=true,
    citecolor=Green,
    linkcolor=NavyBlue,
    filecolor=OrangeRed,      
    urlcolor=OrangeRed,
    pdfpagemode=FullScreen,
}
\newif\ifshowrevs
\showrevstrue

\newif\ifshowauthorcomment
\showauthorcommenttrue

\newif\ifkeep
\keepfalse

\newif\ifremvspace
\remvspacetrue

\newcommand{\idest}{{\it i.e.}, }
\newcommand{\exempli}{{\it e.g.}, }

\newcommand{\wrt}{w.r.t. }

\newcommand{\etc}{{\it etc}}

\newcommand{\sorts}{{SoRTS}}
\newcommand{\trajair}{{TrajAir}}
\newcommand{\xplaneros}{{X-PlaneROS}}

\newcommand{\rosplane}{{ROS-Plane}}
\newcommand{\xplane}{{X-Plane}}

\newcommand{\state}{{\mathbf s}}

\newcommand{\action}{{\mathbf a}}
\newcommand{\R}{{\mathbb R}}
\newcommand{\setS}{{\mathcal{S}}}
\newcommand{\A}{{\mathcal{A}}}
\newcommand{\D}{{\mathcal{D}}}

\newcommand{\modref}{{\bf \textcolor{Orchid!80}{Reference Module}}}
\newcommand{\modval}{{\bf \textcolor{Blue!80}{Cost Map}}}
\newcommand{\modsoc}{{\bf \textcolor{OrangeRed!80}{Social Module}}}

\newcommand{\alghuman}{{\textcolor{Melon}{Human}}}
\newcommand{\algablation}{{\textcolor{Green}{Ablation}}}
\newcommand{\algsorts}{{\textcolor{RoyalBlue}{SoRTS}}}
\newcommand{\user}{{\textcolor{Magenta}{User}}}

\newcommand{\ina}[1]{
    \ifshowauthorcomment
    \textcolor{Green}{[\textbf{Ingrid:} #1]}
    \fi
}

\ifshowrevs

\else

\fi

\usepackage{xargs} 
\usepackage[colorinlistoftodos,prependcaption,textsize=tiny]{todonotes}
\newcommandx{\unsure}[2][1=]{\todo[linecolor=red,backgroundcolor=red!25,bordercolor=red,#1]{#2}}
\newcommandx{\change}[2][1=]{\todo[linecolor=blue,backgroundcolor=blue!25,bordercolor=blue,#1]{#2}}
\newcommandx{\info}[2][1=]{\todo[linecolor=OliveGreen,backgroundcolor=OliveGreen!25,bordercolor=OliveGreen,#1]{#2}}
\newcommandx{\improvement}[2][1=]{\todo[linecolor=Plum,backgroundcolor=Plum!25,bordercolor=Plum,#1]{#2}}
\newcommandx{\thiswillnotshow}[2][1=]{\todo[disable,#1]{#2}}
 
\def\HiLiYellow{\leavevmode\rlap{\hbox to \hsize{\color{yellow!20}\leaders\hrule height .8\baselineskip depth .4ex\hfill}}}
  
\usepackage{cleveref}

\title{\LARGE \bf SoRTS: Learned Tree Search for Long Horizon Social Robot Navigation}

\author{Ingrid Navarro$^{1 \ast}$, Jay Patrikar$^{1 \ast}$, Joao P. A. Dantas$^{1}$, Rohan Baijal$^{1}$, Ian Higgins$^{1}$, \\ Sebastian Scherer$^{1}$ and Jean Oh$^{1}$ \\[-0.8cm]
\thanks{Manuscript received: September, 15, 2023; Revised December, 19, 2023; Accepted February, 5, 2024.}
\thanks{This paper was recommended for publication by Editor Gentiane Venture upon evaluation of the Associate Editor and Reviewers' comments. 
} 
\thanks{$^{1}$Authors are with the Robotics Institute, Carnegie Mellon University, Pittsburgh, PA, USA. {\tt\scriptsize \{ingridn, jaypat, jdantas, ihiggins, basti, jeanoh\}@cs.cmu.edu}}
\thanks{$^{\ast}$Equal contribution.} 
\thanks{Digital Object Identifier (DOI): see top of this page.}
}

\begin{document}

\markboth{IEEE Robotics and Automation Letters. Preprint Version. Accepted February, 2024}
{Navarro \MakeLowercase{\textit{et al.}}: Social Robot Tree Search} 

\maketitle

\begin{abstract}
The fast-growing demand for fully autonomous robots in shared spaces calls for developing trustworthy agents that can safely and seamlessly navigate crowded environments. Recent models for motion prediction show promise in characterizing social interactions in such environments. However, using them for downstream navigation can lead to unsafe behavior due to their myopic decision-making. Prompted by this, we propose \textit{Social Robot Tree Search} (\sorts), an algorithm for safe robot navigation in social domains. \sorts~aims to augment existing socially aware motion prediction models for long-horizon navigation using Monte Carlo Tree Search. 

We use social navigation in \textit{general aviation }as a case study to evaluate our approach and further the research in full-scale aerial autonomy. In doing so, we introduce \xplaneros, a high-fidelity aerial simulator that enables human-robot interaction. We use \xplaneros~to conduct a first-of-its-kind user study where 26 FAA-certified pilots interact with a human pilot, our algorithm, and its ablation. Our results, supported by statistical evidence, show that \sorts~exhibits comparable performance to competent human pilots, significantly outperforming its ablation. Finally, we complement these results with a broad set of self-play experiments to showcase our algorithm's performance in scenarios with increasing complexity. [\href{https://github.com/cmubig/sorts}{Code} $\mid$ \href{https://github.com/castacks/xplane_ros}{Simulator} $\mid$ \href{https://youtu.be/PBE3O4cW2rI}{Video}]
\end{abstract}


\section{Introduction} \label{sec:introduction}

\IEEEPARstart{A}{} social robot strives to synthesize decision policies that enable it to interact with humans, ensuring social compliance while attaining its desired goal. While marked progress has been made in the fields of social navigation \cite{mavrogiannis2021core, rios2015proxemics, tian2022safety} and \textit{socially-aware} motion prediction \cite{rudenko2020human}, achieving seamless navigation among humans while balancing social and self-interested objectives remains challenging.

Classical model-based approaches for social navigation in pedestrian settings have been proposed and remain prominent baselines \cite{helbing1995social, berg2011reciprocal, mavrogiannis2022social}. Yet, their extension to \textit{other domains} is often nontrivial owing to higher environmental and social complexities intrinsic to each domain \cite{wang2022social}. Deep Reinforcement Learning methods have also been vastly explored within the field. Under this formulation, common strategies include approximations to the policy's reward function via hand-crafted design \cite{sadigh2018planning, liu2021decentralized}, or self-play \cite{riviere2021neural, matsuzaki2022learning}. Such techniques are generally promising in settings where data is sparse or where the robot is easily distinguishable from humans. However, tuning reward parameters for homogeneous navigation among humans is challenging \cite{mavrogiannis2017socially}. Furthermore, RL-based policies are a function of the underlying simulator, often yielding undesirable behavior in real scenarios due to a lack of compatibility between the simulated environment and the real world. More recently, data-driven approaches have been extensively explored largely within the field of socially-aware motion prediction. These methods aim at characterizing human behavior and interactions observed in the data, alleviating the need for reward shaping and simulation. These models have achieved promising performance \cite{rudenko2020human}. 
However, using them for downstream navigation can lead to unsafe behavior due to myopic decision-making, prompting the need for robustifying models deployed in the real world.

\begin{figure}[t]
    \centering
    \includegraphics[width=0.45\textwidth,trim={0cm 0cm -0cm -0cm},clip]{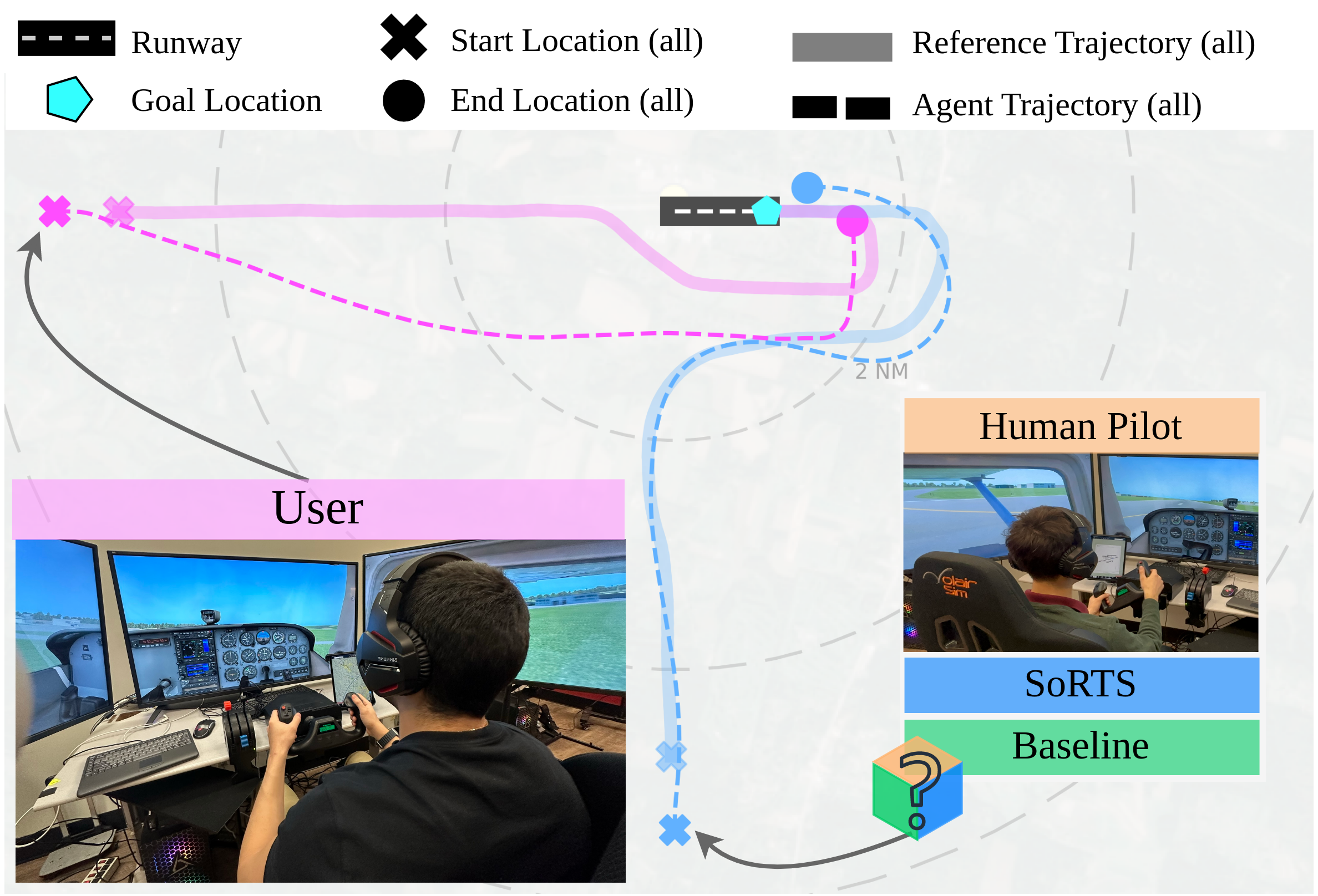}
    \caption{Our flight simulator setup and a user study experiment. In each, the \user~interacts with a \alghuman~pilot, our proposed algorithm, \algsorts, and our \algablation. The figure also shows a resulting interaction between a \user~and \algsorts.}
    \ifremvspace
    \vspace{-0.5cm}
    \fi
    \label{fig:user_study_setup}
\end{figure}

Motivated by these limitations, we introduce \textit{Social Robot Tree Search (\sorts)}, an algorithm that seeks to robustify offline-learned socially aware motion prediction models for their deployment in online settings. We build \sorts~upon the insight that, in social navigation settings, the actions of one agent influence those of another, and vice-versa \cite{sadigh2018planning, hu2022active}. This type of temporally recursive decision-making has been used for modeling human-like gameplay via search-based policies \cite{silver2018general, jacob2022modeling}. We, thus, propose using recursive search-based policies as a means to \textit{augment} the aforesaid prediction methods. Focusing on aleatoric uncertainty, we choose to evaluate and robustify the in-domain performance of prediction models for navigation tasks. Specifically: we use Monte Carlo Tree Search (MCTS) \cite{kocsis2006bandit} as our search policy that provides long-horizon simulations, collision checking, and goal conditioning. 
We then bias its tree search using a \textit{reference} module, which guides the planner to favorable regions, and a \textit{social} module---consisting of the offline-trained motion prediction model---to provide short-horizon agent-to-agent context cues.

We note that \sorts~is intended to be \textit{domain-independent}. Thus, we maintain the formulation details abstract to allow for flexibility in the representation. In this work, however, we use \textit{General Aviation (GA)} as a case study and describe our motivation next. Unmanned Aerial Vehicle (UAV) operations have seen a significant increase in recent years. As a result, there is a growing demand for the development of automated technologies that can be reliably integrated into the National Airspace System \cite{day2021advancing, aurambout2019last, grote2022sharing}, enabling the concurrent use of airspace between human and autonomous aircraft. The next generation of UAVs is expected to operate in low-altitude terminal airspace, where most close-proximity interactions occur \cite{patrikar2022predicting}. Thus, it is crucial to develop technologies that ensure the safe and seamless interaction of these autonomous systems with humans.

Toward this end, recent works have studied the domain of GA in non-towered airports in the context of socially aware motion prediction \cite{patrikar2022predicting, navarro2022socialpattern}, where the goal is to characterize joint interactions in such environments. In this domain, there is no central air traffic control authority to regulate aircraft operations, making pilots solely responsible for coordinating with each other and for following guidelines and traffic patterns established by the FAA to ensure proper operations. Our work builds upon \cite{patrikar2022predicting, navarro2022socialpattern}, exploring this novel domain in the context of social robot navigation. GA is a safety-critical domain demanding the development of competent and trustworthy robots that can properly follow navigation norms but also understand social cues to guarantee \textit{seamless} and \textit{safe} interactions. We separate these two notions into two axes: \textit{navigation efficiency} and \textit{safety}. We center the design of our algorithm, as well as our evaluations around these axes. In doing so, we conducted a user study with 26 experienced pilots using our custom simulator framework, \xplaneros. In our study, we investigate how pilots interact with our model in a realistic flight setting. We also study their perceptions of our model's performance along the preceding axes when compared to competent human pilots. Finally, as a complementary analysis, we provide evaluations via self-play experiments in more complex scenarios. 
\noindent\textit{Statement of Contributions:} 
\begin{enumerate}
    \item We introduce \sorts, an MCTS-based algorithm that aims at robustifying offline-learned socially-aware motion prediction policies for downstream long-horizon navigation. 
    \item We introduce \xplaneros, a high-fidelity simulation environment for navigation in shared aerial space.
    \item We showcase the efficacy of SoRTS in the General Aviation domain. Through a first-of-its-kind user study with 26 FAA-certified pilots and self-play experiments, we show that \sorts~is perceived comparably to a competent human pilot in terms of \textit{navigation efficiency} and \textit{safety} while outperforming its ablation algorithm. 
\end{enumerate}


\section{Related Work} \label{sec:related_works}

\subsection{Social Navigation Methods} \label{ssec:socnav_algos}

Social navigation has a rich body of work primarily focused on human crowds and autonomous driving \cite{mavrogiannis2021core, rudenko2020human}. Several prominent model-based approaches \cite{helbing1995social, berg2011reciprocal, mavrogiannis2022social} have been proposed for pedestrian settings, but their extension to settings with increasing complexities is difficult. RL-based methods \cite{matsuzaki2022learning, chen2017decentralized, chen2019crowd} have produced promising results in these settings by leveraging safety-based handcrafted reward functions. However, shortcomings in simulator design \cite{biswas2022socnavbench}, and domain-specific reward functions limit real-world performance \cite{tsai2020generative}. Achieving scalability and robustness is challenging, often requiring expensive retraining for non-significant test distribution shifts. 

Data-driven approaches focus on exploiting natural behavior by learning policies from datasets that record interactions between agents \cite{tsai2020generative, mavrogiannis2021core}. These models do not need explicit reward construction and, therefore, can capture the rich, joint dynamics of social interactions. However, these methods are challenging to deploy owing to noisy demonstrations, and covariate shifts \cite{bashiri2021distributionally, codevilla2019exploring}. To alleviate this, \cite{sadigh2018planning} used the gradients of a Q-value function for Model Predictive Control, and \cite{hu2022active} proposed a generalization to this method using dual control for belief state propagation. These methods rely on Inverse Reinforcement Learning as an additional step to generate the Q-value functions. Using gradients from sequence models directly in optimizations has also been proposed \cite{schaefer2021leveraging}, but the convergence properties were not examined. Our method differs from the literature in that it does not require specific reward function design or simulator training. Our work is also more direct and intuitive in its use of sequential models, where calculating gradients or Q-values is not required. Instead, we transform the model's outputs into action distributions for the downstream planning task. 

\subsection{Monte Carlo Tree Search for Robot Navigation} \label{ssec:socnav_mcts}

Recent works have leveraged MCTS robot navigation applications. To perform adaptive planning, \cite{ruckin2022adaptive} biases MCTS using a CNN trained to learn informative data gathering actions for fast online re-planning. Their work, however, was not extended for use in dynamic \textit{social} settings. In contrast, \cite{eiffert2020path} proposes a local planner for socially aware settings that leverages MCTS and RNNs to simulate future states. Their predictive model, however, does not account for agent-to-agent interactions, making it unable to characterize rich social dynamics and use them to bias the tree search. Similar to our approach, \cite{riviere2021neural} and \cite{oh2023scan} introduce an RL method that leverages MCTS to train and deploy policies using pre-defined reward functions and simulator training. 

While these methods rely on pre-defined reward functions and simulator training, our work extends them to use offline expert-based policies. Finally, although we believe that the core insights of our work apply to other domains, we provide a domain-specific treatment for social navigation in shared airspace to further the research in this domain.

\ifkeep{
\subsection{Social Navigation Evaluation} \label{ssec:socnav_ev}

\ina{Maybe we can remove this subsection? Set 'keepfalse' to remove it. }
Different metrics have been considered for the evaluation of social robot navigation \cite{mavrogiannis2021core, gao2021evaluation, rudenko2020human}. Some of the main axes of analysis for evaluation include behavior naturalness based on a reference trajectory or irregularity of the executed path \cite{mavrogiannis2019effects, sathyamoorthy2020densecavoid}, performance and efficiency \cite{liu2021decentralized, everett2018motion, liang2020realtime}, and notions of physical personal space or discomfort \cite{torta2013design, chen2017socially}. User studies are also often conducted to evaluate more subjective aspects such as the perceived discomfort, stress and trust that a robot induces during an interaction \cite{butler2001psychological}. Following prior works, we focus on \textit{navigation performance} to measure our agent's smoothness and ability to follow navigation guidelines, and \textit{safety} to judge its ability to respect others' personal space. We also conduct a user study where we ask experienced pilots to interact with our algorithm in a realistic flight setting and rate the robot's performance, perceived safety and trust. 
}
\fi

\begin{figure*}[t!]
    \centering
    \includegraphics[width=0.95\textwidth, trim={0cm 0cm 0cm -0.2cm}]{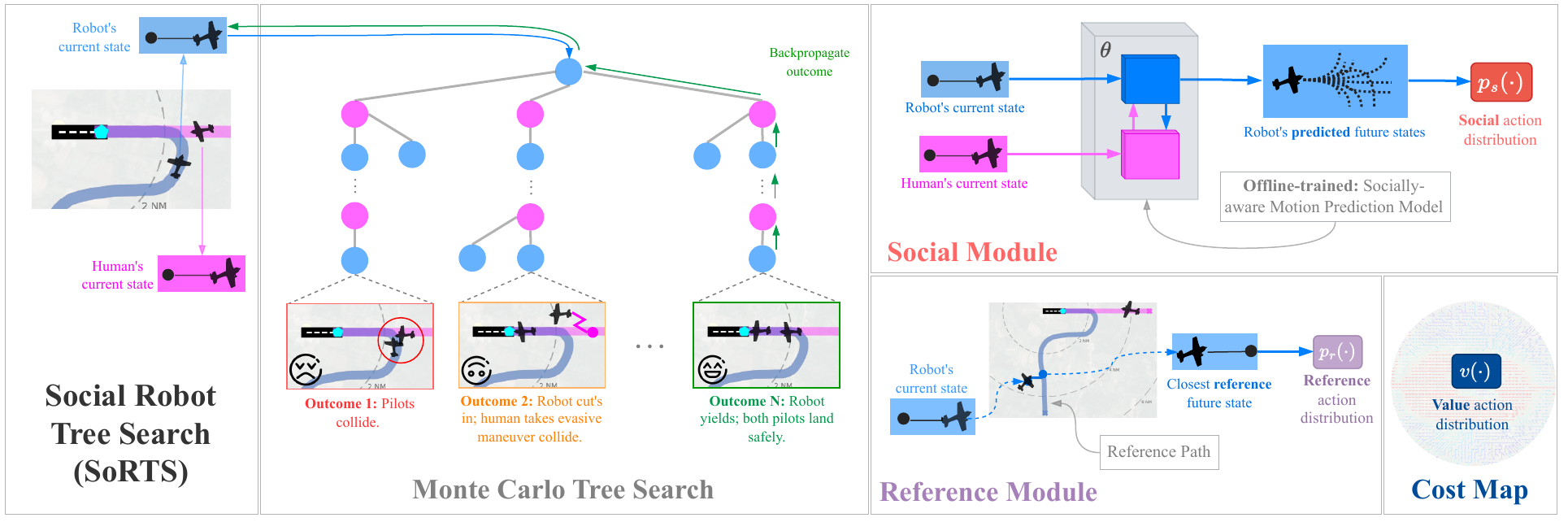}
    \caption{An overview of \sorts, a Monte Carlo Tree Search (MCTS)-based planner for social robot navigation which provides long-horizon simulations, collision checking and goal conditioning. Its tree search is biased by three components; a \modsoc, a \modref~and a \modval. The social module uses a socially-aware motion prediction model to predict a set of possible future states given the \textit{social dynamics} of the scene. The reference module provides the agent with a global path that embodies navigation guidelines the agent must follow. The cost map encodes a global visitation to encourage the agent to move toward more desirable regions.}
    \label{fig:model}%
    \ifremvspace
    \vspace{-0.6cm}
    \fi
\end{figure*}

\section{Problem Formulation} \label{sec:problem_formulation}

\subsection{Notation Details} \label{ssec:notation_details}

We use $\state_t^i \in \setS$ to represent the state of an agent $i$ at time-step $t$, and the agent's start and goal locations as $\state_0^i, \state_g^i$, respectively. Let $\state^i_{t_m:t_n}$ represent a trajectory of states from time-step $t_m$ to time-step $t_n$. Following this, we refer to the observed trajectory of an agent as $\state^i_{t-H:t}$, where $t$ is the current time-step and $H$ is the length of the agent's observed history. We omit superscripts to represent the joint state for multiple agents, \exempli $\state_t = [\state^1_t, \dots, \state^N_t]$. We follow a similar notation for the action, $\action^i_t \in \A$, and a joint action $\action_t$. We use $\tau^i_R = [\state^i_0, \dots, \state^i_g] \in \D$ to represent an agent's reference path. In our work, reference action probabilities are obtained from a dataset, $\D$, comprised of real-world interaction data recorded offline. 
Note that the mathematical details are abstract to allow flexibility in the problem formulation and domain. For our specific representation, refer to \Cref{ssec:implementation_details}.

\subsection{Problem Statement} \label{ssec:definition}

Let us consider a robot (ego-agent) $e$ in a crowded scene. The robot follows a dynamic model, s.t. $\state^e_{t+1} = f(\state^e_t, \action^e_t)$. We assume that the robot can observe the current states of all other agents in the scene via some observation function, s.t. $\state^j_t = o(\cdot), \forall j \neq e$. The goal for the robot is to find a sequence of control inputs $\pi^e = [\action^e_0, \dots, \action^e_g ]$, s.t. it follows a socially-compliant path toward its goal, $\tau^e_S = \{\state_0^e, \dots, \state^e_g\}$. In our work, a socially-compliant path must ensure \textit{safety} and \textit{navigation efficiency}. To satisfy the safety requirement with all agents in the scene, the robot must ensure $||\state^e_t - \state_t^j|| < d, \forall j \neq e$, where $d$ is the minimum separation distance to satisfy the safety objective with all agents in the scene. To satisfy the efficiency component, it needs to stay close to the reference trajectory $\min ||\tau^e_S - \tau^e_R||$.

\section{Approach} \label{sec:approach}

\subsection{An Overview of \sorts} \label{ssec:overview}

\sorts~is an MCTS-based planner whose tree search is guided by the three modules shown in \Cref{fig:model}: (1) a \textit{Social Module} which handles the short-horizon dynamics in the scene, characterizing social cues and interactions; (2) a \textit{Reference Module} which provides the agent with a reference path representing a standardized navigation guideline; and (3) a \textit{Cost Map} which encodes the value function representing the desirability of each state in the state-space. MCTS further provides collision checking and long-horizon simulations. 

The core insight of \sorts~is to use the aforesaid modules to bias MCTS to search through multiple decision modalities, favoring those that prioritize socially compliant and safe behaviors. As a \textit{toy} example, \Cref{fig:model} depicts a situation where the intended paths of two aircraft merge onto a single one, akin to a highway merger. To resolve this situation, the search algorithm will run forward simulations by combining actions. Through them, it will not only prune branches that lead to future collisions (\exempli outcome 1 in the figure) but it will also leverage the social module to choose between socially undesirable (\exempli cutting in, outcome 2) and socially desirable (\exempli yielding, outcome 3) modalities, thereby producing socially-compliant and safe behavior.

\subsection{Modules} \label{ssec:modules}

We provide an overview of the three components used to bias the search algorithm. For specific implementation details on each of these modules, we refer the reader to \Cref{ssec:implementation_details}.

\subsubsection{Social Module} \label{ssec:social} 

To account for the short-term agent-to-agent interactions in a scene, we leverage an offline-learned socially-aware motion prediction model. For an agent $i$ in the scene, the model produces a joint distribution of future actions and states conditioned on the motion histories of all agents in a scene $\state_{t-H:t}$, demonstration data $\D$, and the agent's goal $\state_g^i$, 
\begin{equation}
    p_s(\state_t^i, \action_t^i) \sim P_\theta(\state_t^i, \action_t^i \mid \state_{t-H:t}, \state_g^i, \D)
    \label{soc_action}
\end{equation}  
where $P_\theta$ is the motion prediction model and $\theta$ its learned parameters. While the output of the trajectory prediction model is typically in continuous state space, we convert it into a distribution in discrete motion primitive action space.


\subsubsection{Reference Module} \label{ssec:reference_path}
To encourage the search to adhere to navigation norms, this module calculates the \textit{reference} joint action and state distribution $p_r(\cdot)$ conditioned on reference trajectory $\tau^i_{R} \in \D$.
\begin{equation}
    {p}_r(\state_t^i, \action_t^i) \sim P_r(\state_t^i, \action_t^i | \tau_R^i )
    \label{ref_action}
\end{equation}  
More generally, reference action probabilities can be obtained through other methods, \exempli agent-agnostic global path planners, STL specifications \cite{aloor2022follow}, \etc.


\subsubsection{Cost Map} \label{sssec:cost_map}

To bias the search toward more desirable regions, our algorithm uses a cost map of the environment. The cost map captures the underlying \textit{value} $v(\cdot)$ of the joint state distribution, thus providing a score at a given joint position. The higher the score, the more likely the agents are navigating in favorable regions. These evaluations are used to update MCTS' action value $Q$ during the back-propagation process. In general, $v(\cdot)$ can be learned via self-play \cite{riviere2021neural}, or manually computed from a prior, as said in \Cref{ssec:implementation_details}.

\subsection{Social Monte Carlo Tree Search} \label{ssec:planner}

MCTS is a search-based algorithm frequently used to solve problems requiring sequential decision-making. It expands its tree search toward highly rewarding trajectories by searching the state space and building statistical evidence for the most available decision modalities at a given state \cite{swiechowski2023monte}. MCTS is divided into four stages: selection, expansion, simulation, and back-propagation. During the \textit{selection} process, it uses a tree policy to search the regions of the tree that have been already explored. Most commonly, said policy builds upon the Upper Confidence Bounds applied to Trees (UCT) algorithm \cite{kocsis2006bandit} which proposes a formula that aims to balance the degree of state exploration and exploitation.
\begin{align}
    U(\state, \action) &=  Q(\state, \action) + c \frac{\sqrt{N(\state)}}{N(\state, \action)}
    \label{eq:uct}
\end{align}
where $Q(\state, \action)$ is the empirical average of playing an action $\action$ from state $\state$, $N(\state)$ is the number of times $\state$ has been visited in previous iterations, $N(\state, \action)$ is the number of times $\action$ has been sampled at $\state$, and $c$ is for controlling the exploration and exploitation. Note that we drop the corresponding superscripts and subscripts for ease of notation. In our work, we modify the UCT formula in \Cref{eq:uct} to include the influences from the components presented in \Cref{ssec:modules}, 
\begin{align}
    U(\state, \action) &= 
    Q(\state, \action) + c_1 P_S(\state, \action) + c_2 P_R(\state, \action) \label{eq:social_uct} 
\end{align}
where, $P_S(\state, \action)$, is the normalized visitation component for choosing $\action$ from $\state$ according to the socially-aware prediction network; $P_R(\state, \action)$ is the expected value according to the reference path, and $c_1$ and $c_2$ are the exploration hyper-parameters. These terms are updated according to \Cref{eq:Q}-\ref{eq:PS}.
These updates are performed iteratively within a time budget, or until a leaf state is found. The algorithm further \textit{expands} the tree by adding new leaf states. Then, at each time-step, a new forward \textit{simulation} tree is iteratively constructed by alternately expanding the agents' future states in a round-robin fashion 
{In practice, for $N>2$, we only use the robot and the closest agent to the robot for tree expansion. While the tree is explicitly constructed only for two agents, $p_s$ provides the high-level social context for all the agents. This approximation preserves the real-time nature of the algorithm and is shown to perform well in practice.}. Therein, branches that lead to a collision state are pruned. Upon finalizing this process, the outcomes from the simulations are \textit{back-propagated} to all nodes along the path from the leaf node and the root. Finally, the ego agent uses the updated statistics to select the action that maximizes the normalized visitation count as in \cite{silver2018general}. 
\begin{align}
    Q(\state, \action) &= \frac{N(\state, \action) Q(\state, \action) + v(\state) }{N(\state, \action) + 1} \label{eq:Q} \\[0.5em]
    P_R(\state, \action) &= \frac{N(\state, \action) \cdot P_R(\state, \action) + p_r(\state) }{N(\state, \action) + 1} \label{eq:PR}\\[0.5em]
    P_{S}(\state, \action) &= \frac{\sqrt{N(\state)}}{N(\state, \action)+1} \cdot p_s(\state, \action) \label{eq:PS}
\end{align}

\section{Experimental Setup} \label{sec:setup}



\subsection{Experiment Design} \label{ssec:design}

Our experimental setup involves two or more pilots attempting to land on the same runway at a non-towered airport. The pilots are spawned at the same distance and altitude from the runway. Since there's no central authority to manage the landing, social coordination between aircraft is essential for a safe landing. Moreover, landing at non-towered involves following explicit guidelines established by the FAA \cite{federal2023aeronautical} as well as social norms that exist implicitly in operations at these airports. 
Both, our user study and self-play experiments follow the design outlined above. We provide more details in sections \ref{ssec:evaluation_user_study} and \ref{ssec:self_play_setup}, respectively.  

\subsection{Ablation} \label{ssec:ablation}

Our ablation serves to contrast the benefit of our long-horizon planning strategy against classical social navigation approaches that leverage short-term strategies \cite{mavrogiannis2022social}. We hypothesize that single-step planning leads to worse outcomes as the planner may choose what, in the short-term, seems like the best action to take which in the long-term may not necessarily lead to the best outcome. Thus, the ablation is designed to select the optimal action for each time step by weighting the efficiency and safety components in \Cref{ssec:modules},
\begin{align*}
    \action^\ast = \arg\max_{\action \in \mathcal{A}} \big[ \lambda \cdot p_r(\state_t, \action) + (1 - \lambda) \cdot p_s(\state_t, \action)\big]
\end{align*}
This decision-making strategy loosely follows from \cite{mavrogiannis2022social}, where an agent chooses its next action as a compromise between two objectives. Here $\lambda \in \R: [0, 1]$ is a hyperparameter controlling the importance of the action probabilities given by each objective. 

\subsection{Algorithm Implementation Details} \label{ssec:implementation_details}

We leverage \trajair~\cite{patrikar2022predicting}, a dataset consisting of 111 days of aircraft trajectory data collected in non-towered terminal airspace at the Pittsburgh-Butler Regional Airport, in which standard traffic patterns are followed \cite{airplane}. Our social module was trained offline on \trajair~following the implementation details in \cite{navarro2022socialpattern}. We also used \trajair~to build a library of FAA-abiding global paths used by the reference module. Finally, similar to \cite{aloor2022follow}, we build our cost map based on the flight frequency in \trajair; where we first split the data by discretizing based on the wind direction, and then build a 3D histogram discretizing by aircraft position.

Following \cite{aloor2022follow}, \sorts' state space represents the continuous 3D locations of the aircraft. Its action space consists of 252 motion primitives discretizing the continuous action space in airspeed, vertical speed, and turn angle applied for a length of 20 seconds, following the dynamic model in \cite{aloor2022follow}. Empirically, we set the number of tree expansions and maximum steps to 50 and 100, respectively. Similarly, the exploration parameters in the UCT equation are set to $c_1=2$ and $c_2=5$, and we chose $\lambda=0.3$ for the ablation planner. For further details regarding the state representation, the aircraft dynamic model, and the cost map, we refer the reader to \cite{navarro2022socialpattern, patrikar2022predicting, aloor2022follow}.

\subsection{Simulator} \label{ssec:simulator}

To evaluate our algorithm in our user study, we introduce \xplaneros, a simulation environment that aims at enabling research for human-AI interaction in full-scale aerial autonomy applications. \xplaneros~is a system that combines two main modules; \xplane-11 and \rosplane~autopilot \cite{ellingson2017rosplane}. \xplane-11 \cite{xplane11} is a high-fidelity simulator that provides realistic aircraft models and visuals, as well as an open API that supports multi-agent gameplay. \rosplane~is a widely accepted tool for research and teaching with reliable and autonomous flight control loops. Together, these two modules enable the use of high- and low-level control commands for GA aircraft in realistic world scenarios. \xplaneros~interfaces with X-Plane-11 using NASA’s \xplane Connect \cite{xplaneconnect}. The state information from \xplane~is published over ROS topics. The \rosplane~integration then uses this information to generate actuator commands based on higher-level input to the autopilot system. These actuator commands are then sent back to \xplane~through \xplane Connect. 

\ifkeep{The high-level components of our simulator are depicted in \Cref{fig:sim}
\begin{figure}[t]
    \centering
    \includegraphics[width=\columnwidth]{figures/xplaneros.png}
    \caption{\ina{Maybe we can remove this to save space? Set 'keepfalse' to remove it. }System diagram of \xplaneros~ showing the various components and their interaction. \xplaneros~ is a high-fidelity aerial simulator that combines \xplane~with \rosplane~autopilot. }
    \label{fig:sim}
    \ifremvspace
    \vspace{-0.5cm}
    \fi
\end{figure}
\fi

\section{Experiments}

\subsection{User Study} \label{ssec:evaluation_user_study}

We recruited 26 FAA-certified pilots\footnote{We received approval from Carnegie Mellon University's Institutional Review Board for this study (protocol no. \textit{STUDY2022\_00000195}, approved June 15th, 2022). The committee authorized all aspects, including data handling, and adherence to ethical guidelines. All participants gave written informed consent. We ensured participant anonymity.} (14 private, 8 commercial, 3 student pilots, and 1 airline transport pilot), who on average have 986 flight hours. Each pilot had to complete a set of landing tasks on a specified runway at an airport using \xplaneros~and a flight deck like those shown in \Cref{fig:user_study_setup}. In each task, a second pilot was simultaneously attempting to land on the same runway. Here, the second pilot was either a human, \sorts~or the ablation discussed in \Cref{ssec:ablation}. In the subsequent paragraphs, we use \textit{second pilot} and \textit{algorithm} interchangeably. 

We followed a \textit{within-subject} design where each user tested against each algorithm. We let users get familiarized with the simulator and controls before beginning the tasks. In each instance, we spawn the pilots within a 10 km radius. Their incoming direction was either north (N), south (S) or west (W), defining six possible scenarios for the pilot pair: $\{(N, S), (S, N), (N, W), (W, N), (S, W), (W, S)\}$. The algorithm order and scenario were randomly chosen. The scenario remained fixed throughout the three tests. We note that pilots do not see the reference trajectory, but they have access to a global map in the simulator and are instructed to follow the standard flight patterns.

After each task, users completed a 5-point Likert questionnaire to evaluate the second pilot's performance along our two notions of interest: efficiency and safety. To assess efficiency, we asked users to rate the other pilot's (1) ability to adhere to FAA guidelines, (2) overall flying skill, and (3) flight smoothness. For safety, users rated the second pilot along the following components: (1) collision risk, (2) comfort, (3) abruptness, (4) cooperativeness, and (5) predictability. 
The users were also asked to rate the \textit{trustworthiness} of each algorithm to gain insights into which components of efficiency and safety were deemed more relevant. 
Finally, we also collected the trajectory data from the experiments to further analyze efficiency and safety using the metrics in \Cref{ssec:metrics}.


\subsection{Self-Play} \label{ssec:self_play_setup}

Recruiting certified pilots to conduct our evaluations is challenging. As such, our user study focused on evaluating a limited number of scenarios. To assess \sorts' performance on a broader set of scenarios, we complement our study with self-play simulation experiments. 
Our self-play simulation experiments follow a similar design to that of the proposed user study; \idest agents are spawned 10km away from the airport, and their task is to land at the specified runway. In contrast with the user study, however, the initial location for each agent can be any location around the 10km radius to encourage higher scenario diversity. 

We consider multi-agent scenarios with 2 to 5 agents. We randomly generated 100 episodes for each setting, where, an agent is deemed unsuccessful if it breaches a minimum separation distance with another agent (\Cref{ssec:metrics}), gets off-track, or reaches a maximum number of allowed steps. 

\subsection{Metrics} \label{ssec:metrics}

Following \cite{mavrogiannis2021core, gao2021evaluation}, we also assess the performance of our user study and self-play experiments using the following objective metrics: (1) \textit{Reference Error (RE):} the distance between a reference trajectory and the agent's executed path. 
(2) \textit{Loss of Separation (LS):} the duration that two agents break a minimum distance from each other. Although LS is specific to the domain of aviation \cite{glozman2021vision}, it is akin to \textit{personal space} metrics commonly used within pedestrian settings to assess the level of discomfort incurred by the robot to the surrounding agents \cite{mavrogiannis2021core}. 
We use RE and LS as proxies for efficiency and safety.

\section{Results} \label{sec:results}

\begin{figure*}[!ht]
    \centering
    \includegraphics[width=0.96\textwidth, trim={0cm 0cm 0cm -2cm}]{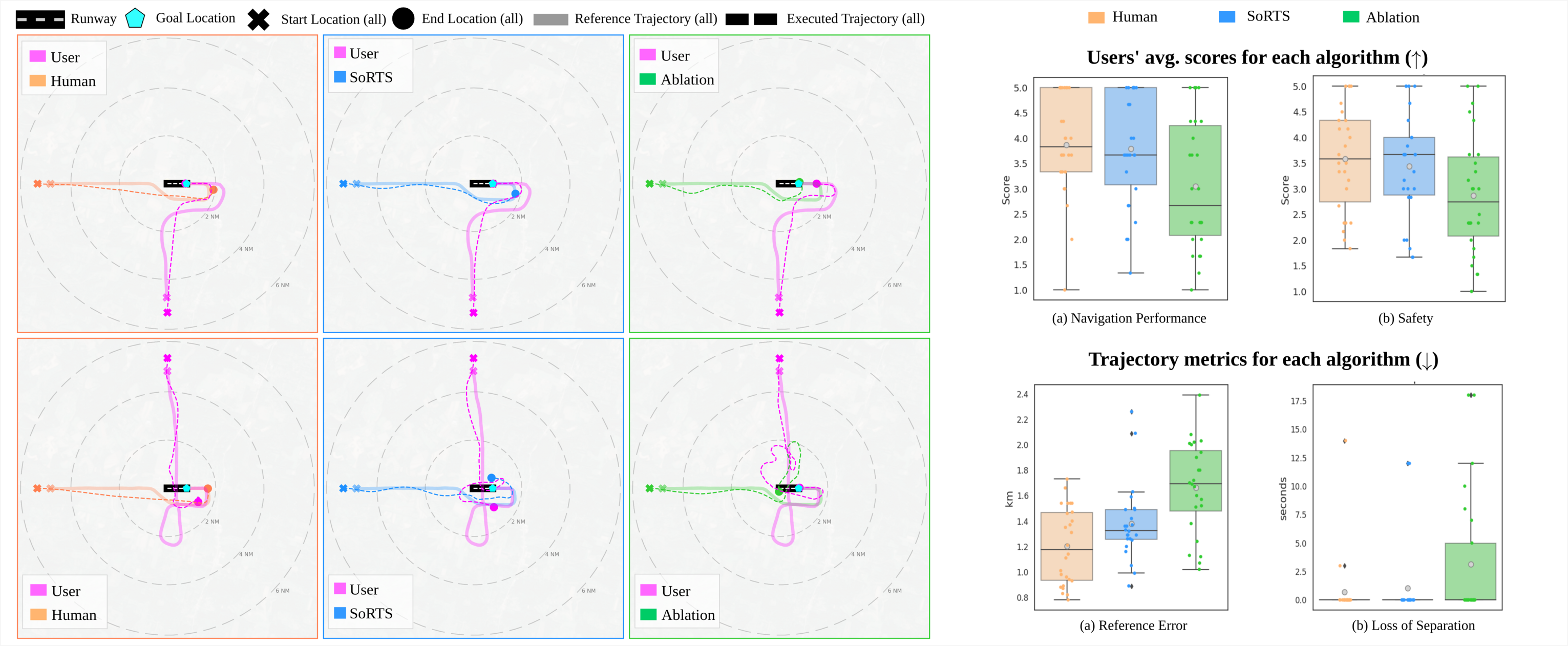}
    \caption{User study results. \textbf{Left:} Each row shows the resulting trajectories of a \user~interacting with our \alghuman~pilot, \algsorts, and the \algablation; reference paths are shown as solid lines, executed ones as dashed lines. The top row shows a user that successfully followed the expected path. We can observe that the ablation did not follow the reference path as smoothly as \sorts, and also cut short when approaching the goal. The bottom row shows a user that did not follow the expected path. Here, \sorts~still managed to navigate properly. In contrast, the ablation behaved erratically, unsafely crossing over the runway twice. \textbf{Right:} Box-plots showing per-algorithm results. The top ones show the avg. efficiency (a) and safety (b) scores given by the users. The bottom ones show the avg. RE (a) and LS (b) metrics obtained from the trajectory data.}
    \label{fig:user_study}
    \ifremvspace
    \vspace{-0.4cm}
    \fi
\end{figure*}

In \Cref{ssec:trust}, we first analyze how human pilots perceived each algorithm \wrt \textit{trustworthiness} based on the performance components in \Cref{ssec:evaluation_user_study}. We then use those results to compare the algorithmic pairs in \Cref{ssec:algorithm_pairs}.  

\subsection{On performance and trust} \label{ssec:trust}

We first compare the factors from \Cref{ssec:evaluation_user_study} with the user's perceived trust for each of the \textit{challenger} pilots. To do so, we perform a Pearson's correlation with repeated measures analysis and summarize the results in \Cref{tab:us_rm_corr}. The table shows that the users' perceived trust strongly correlates to all of the factors within the axis of navigation efficiency, hinting that users notice and prioritize aspects relating to flight smoothness and the ability to follow navigation guidelines. For the safety aspect, we observe that the users' assessments for trust were more strongly correlated to how comfortable the users felt during the interaction and how erratic or unpredictable was the behavior of the pilot. 

\ifkeep
\subsection{On competence and human performance} \label{ssec:humaness}

Similar to the previous section, we also ask users to gauge whether the second pilot was a human based on their responses to the questionnaire in \Cref{ssec:evaluation_user_study}. Surprisingly, we find a marked disagreement within the users' responses to this question, with almost a 50-50\% split between the responses for the \textit{Human Pilot}~(14: No, 12: Yes) and the \textit{ablation}~(14: No, 12: Yes), whereas for {\it\sorts}~(8: No, 18: Yes) more users perceived its performance as human-like. Further highlighting this disagreement, \Cref{tab:us_rm_corr} shows that the user's assessments for navigational performance and safety correlate weak with the \textit{humanness} prediction. Even isolating the responses for \sorts~from the other two, we find weak correlations along our factors of interest, with \textit{predictability} (R=0.53, p=0.01) being the strongest one.  \ina{If enough space - add more discussion, especially regarding why pilots thought of SoRTS as more human like.}
\fi

\subsection{On the performance of each algorithm} \label{ssec:algorithm_pairs}

We now provide a comparison between the \textit{algorithms} in the user study and self-play experiments. 

\subsubsection{User study} 

For each algorithm, we first obtain average scores for the efficiency and safety leveraging the trustworthiness results in \Cref{tab:us_rm_corr}. For the efficiency component we compute the mean score between \textit{following FAA guidelines} and \textit{flight smoothness}, which correspond to the highest correlating factors for this axis. For safety, we use \textit{predictability} and \textit{comfort}. The scores are shown in \Cref{fig:user_study} (top-right). 

\begin{table}[!ht]
\centering
\caption{Pearson correlations (R, p-value=0.05) for each component in \Cref{ssec:evaluation_user_study} \textit{vs.} perceived trustworthiness.}
\label{tab:us_rm_corr}
\resizebox{0.9\columnwidth}{!}{%
\begin{tabular}{@{}clcccc@{}}
\toprule
 \multirow{2}{*}{Axis} & \multirow{2}{*}{Factor} & & & \multicolumn{2}{c}{Trustworthiness} \\
 \cmidrule(l){5-6} 
 & & & & R & p-value \\ \midrule
\multirow{3}{*}{Nav. Efficiency}  & Flight Smoothness & & & 0.81 & 4.80e-19  \\ 
 & Follow FAA Guidelines & & & 0.76 & 2.99e-15  \\
 & Overall Flying Skill & & & 0.71 & 4.56e-13  \\
\midrule
\multirow{4}{*}{Safety}  & Comfortable & & & 0.92 & 3.55e-31  \\
  & Predictable Behavior & & & 0.77  & 5.19e-16  \\
  & Cooperative & & & 0.65 & 1.61e-10 \\
  & Collision Risk & & & -0.56 & 1.10e-07 \\
  & Abrupt     & &  & -0.56 & 1.13e-07 \\
\bottomrule
\end{tabular}
}
\ifremvspace
\fi
\end{table}

We then use ANOVA with repeated measures to compute the pairwise statistical differences between the algorithms shown in \Cref{tab:us_statistical_analysis}. Our analysis suggests that there is no statistical evidence that the scores for Human and \sorts~were different. This hints that the users rated their performances similarly. We also find that the ablation was generally rated lower on both of these axes while also displaying higher variance, compared to the other algorithms. We compute the RE and LS metrics on the resulting trajectories to examine how they tie to the users' assessments for efficiency and safety. Their corresponding average values and statistical analysis are also shown in \Cref{fig:user_study} (bottom-right) and \Cref{tab:us_statistical_analysis}. For the RE metric, we find marked difference between the algorithms; wherein \sorts~yields higher error compared to the human pilot, but lower than the ablation. \sorts~also exhibits less variance than the other two. For the safety metric we observe that, in general, neither the human pilot nor \sorts~breach the safe distance here set to 0.3 km. In contrast, we see that the ablation does it more frequently, creating more situations for potential collisions. 

We also show trajectory visualizations of our experiments in \Cref{fig:user_study}. Each row represents one user \textit{vs.} the three \textit{algorithms}. We show a reference path along with the executed path. The top row shows an example of a successful user. Here, we observed that the ablation unexpectedly cuts short while approaching the runway instead of following the reference path. In contrast, \sorts~yields to the user and smoothly follows the expected pattern. The bottom row shows a user that did not follow the standard navigation pattern. Despite this, \sorts~manages to successfully complete the task, while the ablation behaves erratically, not following the pattern and unsafely traversing the runway twice. 

\begin{table}[!ht]
\centering
\caption{Statistical significance between algorithmic pairs for results in \Cref{fig:user_study} with $t^\ast \geq 2.060$ and $p\leq0.05$.}
\label{tab:us_statistical_analysis}
\resizebox{\columnwidth}{!}{%
\begin{tabular}{@{}cc@{\hspace{0.2cm}}cc@{\hspace{0.2cm}}cc@{\hspace{0.2cm}}cc@{\hspace{0.2cm}}c@{}}
\toprule
\multirow{2}{*}{\textbf{Algorithmic}} & \multicolumn{2}{c}{\textbf{Nav. Efficiency}} & \multicolumn{2}{c}{\textbf{Safety}} & \multicolumn{2}{c}{\textbf{RE}} & \multicolumn{2}{c}{\textbf{LS}} \\ \cmidrule(l){2-9} 
 \textbf{Pair} & t-val & p-val & t-val & p-val & t-val & p-val & t-val & p-val \\ \midrule
Ablation-Human & 3.121 &  0.009 & 3.062 & 0.016 & 5.782 & 0.000 & 1.321 & {\color{OrangeRed} 0.199} \rule{0pt}{2.6ex}\\
Ablation-\sorts & 3.018 &  0.009 & 2.626 &  0.022 & 2.105 & 0.011 & 0.397 & {\color{OrangeRed} 0.694} \rule{0pt}{2.6ex}\\
Human-\sorts & {\color{OrangeRed} 0.415} & {\color{OrangeRed} 0.682} & {\color{OrangeRed} 0.322} & {\color{OrangeRed} 0.322} & 2.944 & 0.022 & 1.211 & {\color{OrangeRed} 0.237}  \rule{0pt}{2.6ex} \\ 
\bottomrule
\end{tabular}
}
\vspace{-0.2cm}
\end{table}

We conclude that \sorts~performs comparable to a competent human pilot and significantly better than the ablation, both, as perceived by the users and according to our metrics. 

\subsubsection{Self-play}

\begin{figure}[!h]
\centering
\includegraphics[width=0.45\textwidth, trim={0cm 0cm 0cm -2cm}]{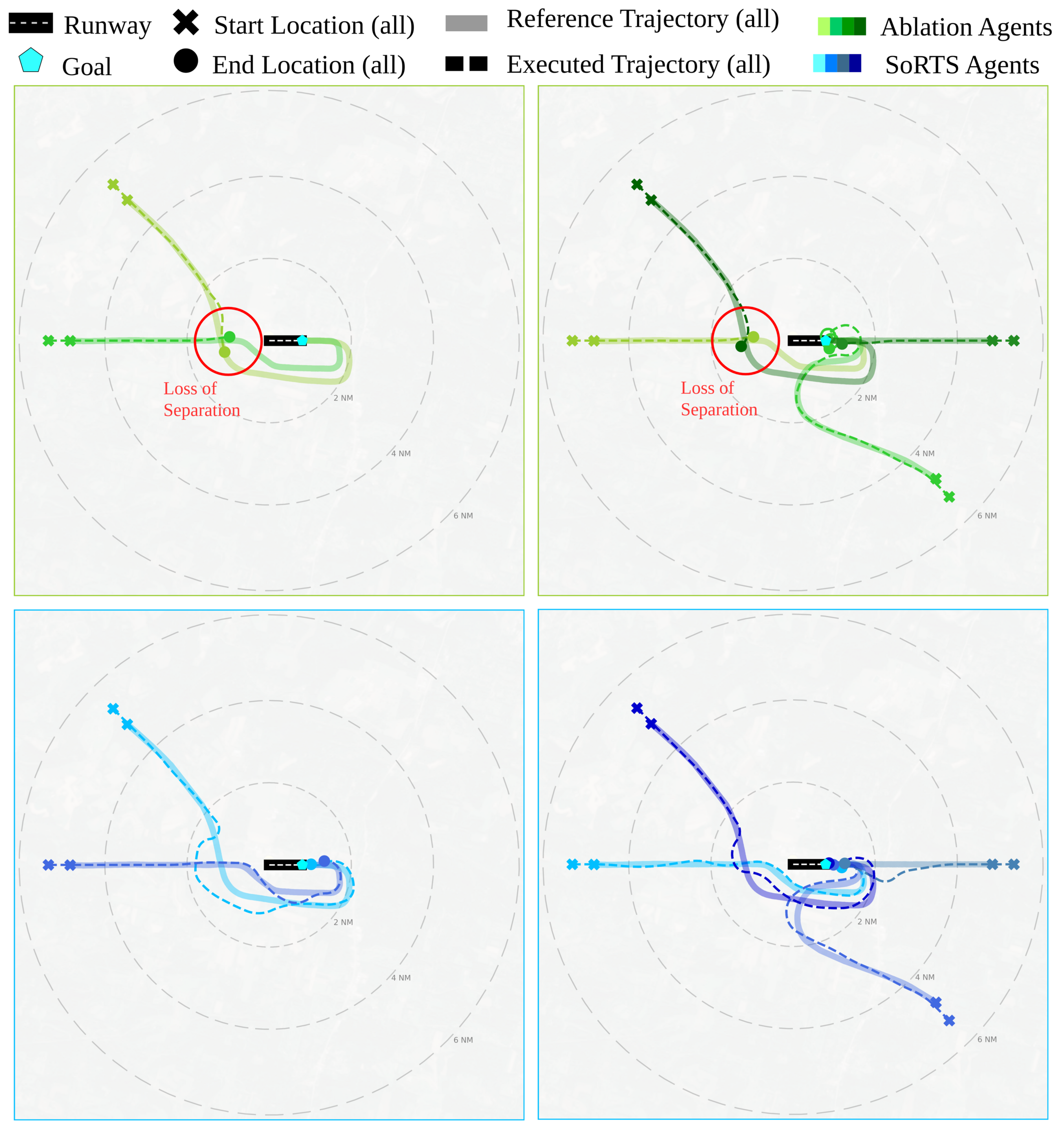}
\caption{Self-play results. The top row shows examples where the \algablation~agents fail to resolve social conflicts and end in a loss of separation situation. With the same initial conditions, the bottom row shows that \algsorts~agents are able to adjust their paths to properly resolve these situations.}
\label{fig:self-play_experiments}
\ifremvspace
\vspace{-0.5cm}
\fi
\end{figure}

We summarize the results of the self-play experiments in \Cref{tab:self-play_results}. The table shows the percentage of \textit{successful} and \textit{unsuccessful} agents and their average RE. Here, task success was higher for \sorts~than for the ablation for all experiments. Although we observe a decrease in task success as the number of agents increases for both algorithms, we see a more significant drop in performance for the ablation ($\sim$21\%) than \sorts~($\sim$11\%) as the number of agents goes from 2 to 5. Finally, for all failure conditions, the failure percentage was significantly lower for \sorts. Finally, \Cref{fig:self-play_experiments} shows qualitative results of the ablation~(top) and \sorts~(bottom) experiments. We show situations in which the ablation agents, unable to resolve social conflicts, fail to complete their task due to a loss of separation. Then, under the same initial conditions, \sorts~agents are able to foresee and avoid the potential conflict and successfully complete their task.

\begin{table}[!th]
\centering
\vspace{0.2cm} 
\caption{Task performance summary for self-play agents.}
\label{tab:self-play_results}
\resizebox{\columnwidth}{!}{%
\begin{tabular}{@{}ccccccc@{}}
\toprule
\multirow{2}{*}{\textbf{\# Agents}} & \multirow{2}{*}{\textbf{Algorithm}} & \multirow{2}{*}{\textbf{Success ($\uparrow$\%)}} & \multicolumn{3}{c}{\textbf{Failure ($\downarrow$\%)}} & \multirow{2}{*}{\textbf{RE}} \\ \cmidrule(l){4-6}
 &  &  & \textbf{LS} & \textbf{Timeout} & \textbf{Offtrack} & \\ 
\midrule
\multirow{2}{*}{2} & Ablation & 88.0 & 12.0 & 0.00 & 0.00 & 2.05  \rule{0pt}{2.6ex} \\
                   & \sorts   & 92.5 &  4.0 & 1.00 & 2.50 & 2.15  \rule{0pt}{2.6ex} \\
\midrule
\multirow{2}{*}{3} & Ablation & 74.7 & 18.0 & 5.30 & 2.00 & 2.02  \rule{0pt}{2.6ex} \\
                   & \sorts   & 86.3 & 12.7 & 0.30 & 0.70 & 2.03  \rule{0pt}{2.6ex} \\
\midrule
\multirow{2}{*}{4} & Ablation & 74.0 & 18.5 & 5.30  & 3.20 & 2.09 \rule{0pt}{2.6ex} \\
                   & \sorts   & 86.0 & 13.0 & 0.75 & 0.25 & 2.07  \rule{0pt}{2.6ex} \\
\midrule
\multirow{2}{*}{5} & Ablation & 69.4 & 21.2 & 6.20 & 3.20 & 2.04  \rule{0pt}{2.6ex} \\
                   & \sorts   & 81.8 & 16.4 & 1.60 & 0.20 & 2.06  \rule{0pt}{2.6ex} \\
\bottomrule
\end{tabular}
}
\begin{tablenotes}
\scriptsize
\item \textbf{LS}: Loss of separation at 0.2km, \textbf{RE}: reference error in km for successful agents. 
\end{tablenotes}
\vspace{-0.2cm} 
\end{table} 

\section{Conclusion} \label{sec:conclusions}


We present \sorts, an MCTS-based planner that aims to augment offline-trained socially aware motion prediction models for their deployment in online social navigation settings. 
Our work uses the domain of general aviation as a use case. In doing that, we introduce \xplaneros, a high-fidelity simulator for research in full-scale aerial autonomy. We use it to conduct a user study with experienced pilots to study our algorithm's performance in realistic flight settings. We find that users perceive \sorts~comparable to a competent human pilot and significantly better than our baseline. In self-play, we show that \sorts~outperforms the baseline by up to 17.9\% on the more crowded and complex scenarios.


We identify two main avenues for future work. 
Firstly, our work assumes \textit{perfect} intent and state estimation. The scope of the paper is restricted to evaluating and robustifying the in-domain performance of prediction models for navigation tasks, thus focusing on aleatoric uncertainty. This challenge is in contrast to robustifying against out-of-distribution scenarios which is a different line of research as detailed in the future work section. Accordingly, robustifying prediction models with uncertainty and adversarial awareness is a promising direction \cite{farid2023task, zhang2022adversarial}. 
Finally, we believe the core insights of our approach and the modularity of its design make it suitable for its application to other domains. 


\section*{ACKNOWLEDGMENT}

\small{This work was supported by the Army Futures Command Artificial Intelligence Integration Center, the Korean Ministry of Trade, Industry and Energy, the Korea Institute of Advancement of Technology, and the Brazilian Air Force. We thank Condor Aero Club and ABC Flying Club pilots for participating in our user study.}

\bstctlcite{IEEEexample:BSTcontrol}
\bibliographystyle{IEEEtran}  
\bibliography{IEEEabrv,ref}

\end{document}